# Guiding Visual Attention in Deep Convolutional Neural Networks Based on Human Eye Movements

**Leonard Elia van Dyck[1,2]\*, Sebastian Jochen Denzler[1], and Walter Roland Gruber[1,2]**

[1]Department of Psychology, University of Salzburg, Austria
[2]Centre for Cognitive Neuroscience, University of Salzburg, Austria

**\*Correspondence:**
Leonard Elia van Dyck
leonard.vandyck@plus.ac.at

**Keywords: vision, attention, brain, deep neural networks, eye tracking, saliency map, object recognition, face detection**

## Abstract

Deep Convolutional Neural Networks (DCNNs) were originally inspired by principles of biological vision, have evolved into best current computational models of object recognition, and consequently indicate strong architectural and functional parallelism with the ventral visual pathway throughout comparisons with neuroimaging and neural time series data. As recent advances in deep learning seem to decrease this similarity, computational neuroscience is challenged to reverse-engineer the biological plausibility to obtain useful models. While previous studies have shown that biologically inspired architectures are able to amplify the human-likeness of the models, in this study, we investigate a purely data-driven approach. We use human eye tracking data to directly modify training examples and thereby guide the models' visual attention during object recognition in natural images either towards or away from the focus of human fixations. We compare and validate different manipulation types (i.e., standard, human-like, and non-human-like attention) through GradCAM saliency maps against human participant eye tracking data. Our results demonstrate that the proposed guided focus manipulation works as intended in the negative direction and non-human-like models focus on significantly dissimilar image parts compared to humans. The observed effects were highly category-specific, enhanced by animacy and face presence, developed only after feedforward processing was completed, and indicated a strong influence on face detection. With this approach, however, no significantly increased human-likeness was found. Possible applications of overt visual attention in DCNNs and further implications for theories of face detection are discussed.

# 1 Introduction

With the groundbreaking discovery of simple and complex cells as well as their receptive field arrangements, Hubel and Wiesel (1959, 1962) not only laid the foundation for decades of subsequent findings in visual neuroscience but also ignited an idea in another field emergent at that time – computer vision. The *Neocognitron* was introduced by Fukushima and Miyake (1982), who translated the aforementioned findings to one of the earliest multi-layer artificial neural networks and thereby set the cornerstone for image and pattern understanding in deep learning. Nearly half a century later, computer and biological vision research cannot be imagined without the Neocognitron's well-known successors, commonly referred to as *Deep Convolutional Neural Networks* (DCNNs; LeCun et al., 2015), the state-of-the-art models of biological processes such as object recognition. View-invariant core object recognition, that is "the ability to rapidly recognize objects despite substantial appearance variation" (DiCarlo et al., 2012), is a key mechanism in visual information processing that allows humans and animals to perceive, understand, and act upon the surrounding environment.

Today, a lot of research is devoted to the architectural and functional similarities of the brain and DCNNs during visual object recognition (Cichy & Kaiser, 2019; Kietzmann, McClure, et al., 2019; Richards et al., 2019; Storrs & Kriegeskorte, 2019; Yamins & DiCarlo, 2016). Evidence has been steadily accumulated throughout different modalities and measurements (i.e., primarily brain imaging and neural time series data). This remarkable body of literature suggests that especially the hierarchical organization of both biological and artificial cascades seems to be highly similar (Cadieu et al., 2014; Cichy et al., 2016; Greene & Hansen, 2018; Güçlü & van Gerven, 2015; Horikawa & Kamitani, 2017; Khaligh-Razavi & Kriegeskorte, 2014; Kheradpisheh et al., 2016; Schrimpf et al., 2020; Yamins et al., 2013; Yamins et al., 2014). Neural activations of areas along the ventral visual pathway, which is known to be primarily specialized in object recognition (Ishai et al., 1999), and network layers can be mapped onto each other in an ascending spatial and temporal fashion. While earlier regions of the visual cortex (e.g., primary visual cortex V1 or secondary visual cortex V2) correlate with predominantly earlier model layers and encode mainly lower-level features such as edges, blobs, and color, later regions (e.g., visual area V4 or inferior temporal cortex ITC) coincide with later model layers in their functionality of grouping conceptually higher-level features in order to recognize parts and in turn entire objects (LeCun et al., 2015; Marr, 1982). Despite this striking parallelism, there is also a lot of research that revolves around the fundamental dissimilarities between the ventral visual pathway and its computational models. Several studies have demonstrated that in addition to feedforward propagation of neural activity, the processing of some visual inputs may require additional recurrent and feedback connections (Felleman & Van Essen, 1991; Lamme & Roelfsema, 2000; Muckli, 2010; Muckli et al., 2015), which are presumably only activated in primary visual areas after an initial end-to-end sweep of ~150-200 ms (Kar & DiCarlo, 2020; Loke et al., 2022; Thorpe et al., 1996) and entirely missing in off-the-shelf DCNNs (Kar et al., 2019; Kietzmann, Spoerer, et al., 2019). This is not surprising, given the fact that the cortex is a heavily specialized, hierarchical architecture with an abundance of differently directed neuronal projections (for review see Bastos et al., 2012).

**Guiding Visual Attention in DCNNs**

Even though DCNNs have already reached human benchmark performance in object recognition tasks (He et al., 2016; Krizhevsky et al., 2012), recent advances in computer vision hold out the prospect of further milestones with increasingly deep and sophisticated models. However, from a neuroscientific perspective, this development introduces DCNNs that operate in increasingly dissimilar fashion compared to the postulated biological mechanisms (Nonaka et al., 2021), which makes them incomparable to human vision and less useful as explanatory models. In contrast, a key mission of vision science is to reverse-engineer object recognition to build better and, in turn, infer from more brain-like models.

Therefore, a question of both scientific and practical nature arises: *How can we develop DCNNs that use more human-like processes to recognize objects?* Theoretically, the most suitable answer to this demand would incorporate both an architecture similar to that of the ventral visual pathway and analogical data that the brain is using to solve this particular challenge. In a previous study, we provided evidence that a DCNN with a receptive field organization resembling that of the ventral visual pathway displays more human-like visual attention in object recognition (van Dyck et al., 2021). With eye tracking, we recorded fixations of human participants during an object categorization task with natural images and compared it to *Gradient-weighted Class Activation Mapping* saliency maps (Selvaraju et al., 2017; hereafter referred to as GradCAMs) of several models with different biological plausibility. The architectural modification yielded more human-like visual attention, as the biologically inspired model focused more on image parts that primarily attracted fixations. Our findings complemented the work by Mehrer et al. (2021), who introduced an image dataset called *Ecoset*, which organizes objects in entry-level categories commonly used by humans (Rosch, 1978). Interestingly, training standard DCNNs on this restructured entry-level category system (e.g., *bird* and *fish*) already improved the model's similarity to human fMRI activity significantly compared to a subordinate category structure (e.g., *great grey owl* and *coho salmon*) – the default in computer vision. This finding leads us to the assumption that a data-driven manipulation of training examples based on human eye movements should likewise increase the similarity between *in vivo* and *in silico* object recognition. Therefore, in this paper, we focus on increasing the human-likeness of visual attention in DCNNs through manipulated visual inputs, rather than previously reported modified architecture, to demonstrate that there are generally two approaches in the sense of nature (i.e., innate mechanisms or architecture) and nurture (i.e., experience or training data) to influence visual attention. To achieve this, we apply a fine-tuning approach to a widely used DCNN called AlexNet (Krizhevsky et al., 2012), in order to guide its focus during visual information processing towards features that are highly relevant to humans and thereby increase its human-likeness in object recognition. In this procedure, we use human observers' fixation density maps (hereafter referred to as eye tracking heatmaps) as a blueprint for a blurring manipulation applied to the training images accordingly. While image parts, that human observers fixated on, remain clearly visible and unchanged in information, parts that did not receive any overt visual attention are blurred and thereby reduced in informational content. Through this *Human-Spotlight* manipulation, the models should additionally be trained to "see" through the lens of and use features meaningful to human observers.

**Guiding Visual Attention in DCNNs**

In this study, we aim to shed light on three sets of research questions. Hard and soft attention mechanisms in deep learning (for review see Niu et al., 2021) have been primarily investigated from an engineering standpoint to solve practical problems (Juefei-Xu et al., 2016). Here, we want to show that this approach could be a key tool to incorporate human-likeness in DCNNs and by doing so take a step forward towards explainable artificial intelligence (Adadi & Berrada, 2018). Therefore, the first section of this paper will revolve around the question (i) *whether guided focus based on human eye movements – the process of not only teaching a model how to correctly categorize but additionally where to look for information relevant to humans – increases the human-likeness of the models*. We will use human eye tracking heatmaps acquired in an object categorization paradigm (see Figure 1A) to blur images accordingly, which are then used for fine-tuning different DCNNs originating from a common pretrained architecture (see Figure 1B). We hypothesize that the model should learn manifest, accuracy-relevant as well as latent, similarity-relevant information and thereby increase in human-likeness. Furthermore, recent evidence from Jang et al. (2021) is important to take into account here, as it suggests that training several state-of-the-art DCNNs, including AlexNet, on images with added noise increases the models' similarity to human behavioral performance and neural representations under challenging conditions. They argue that this might be a result of the increased noise-robustness found in later layers of these models. Therefore, in addition to our *Human-Spotlight* experimental condition (hereafter HS), we incorporate an *Anti-Spotlight* control condition (hereafter AS; see Figure 1C), the inverse of the former manipulation, to show that it is possible to guide visual attention of DCNNs in either direction and that the observed effects are a result of the intended manipulation and not simply due to training on noisy data.

Another important question we want to address in the second section of this paper is (ii) *how the growing body of neural time series and neuroimaging literature on the temporal unfolding of feedforward and feedback visual processing can be tested in the realm of eye tracking*. In line with powerful frameworks such as predictive coding theory (Friston, 2005), feedback connections in the visual cortex are thought to drive primarily top-down modulations of the bottom-up sensory input based on internally generated predictions of the external world. This idea fits well with findings on manipulated images that seem to require especially recurrent/feedback processing after ~150-200 ms of feedforward processing and therefore considerably challenge DCNNs (Kar et al., 2019; Rajaei et al., 2019; Tang et al., 2018). An illustrative example of this are occlusion manipulations, where a part of an image is masked. Based on general assumptions of predictive coding, this covert part should elicit a strong prediction error, as the bottom-up sensory input disagrees with and must be replaced by top-down predictions. As shown by Muckli et al. (2015) with high-resolution layer fMRI, in the case of occlusion, more feedback connections are activated in the cortical layers of V1, as this region seems to receive additional top-down attention. Thus, contrary to parts of our previous analyses (van Dyck et al., 2021), we have come to the general assumption that the initial feedforward sweep has to be completed for preattentive processing before systematic eye movements (i.e., saccades) can be planned and increasingly feedback, attentive processing of critical features (i.e., fixations) can set in (Treisman, 1985; Van der Stigchel et al., 2009). Hence, we hypothesize that the similarity between eye tracking heatmaps and GradCAMs should abruptly increase and peak after ~150-200 ms postulated by neural time series data and decrease continuously, as processing shifts gradually from feedforward to feedback mechanisms and purely feedforward DCNNs should not be able to capture the latter.

The third section of this paper will follow up on (iii) *possible methodological advantages and disadvantages of forcing a model to use specific features.* As previous works reported performance decreases in noise-trained DCNNs when tested on natural, noise-free images (Geirhos et al., 2018; Jang et al., 2021), here we hypothesize that an increase in similarity causes a decrease in accuracy and vice versa. As the model generally optimizes its parameter weights, changing the availability of these features should, in most cases, lead to comparatively suboptimal feature distributions with noise added by the image manipulation. To investigate this trade-off for the HS and AS manipulation, we will fine-tune model versions with different stepwise ratios of natural and HS or AS images and describe how accuracy and similarity change as a function of the magnitude of our guided focus manipulation (see Figure 1D).

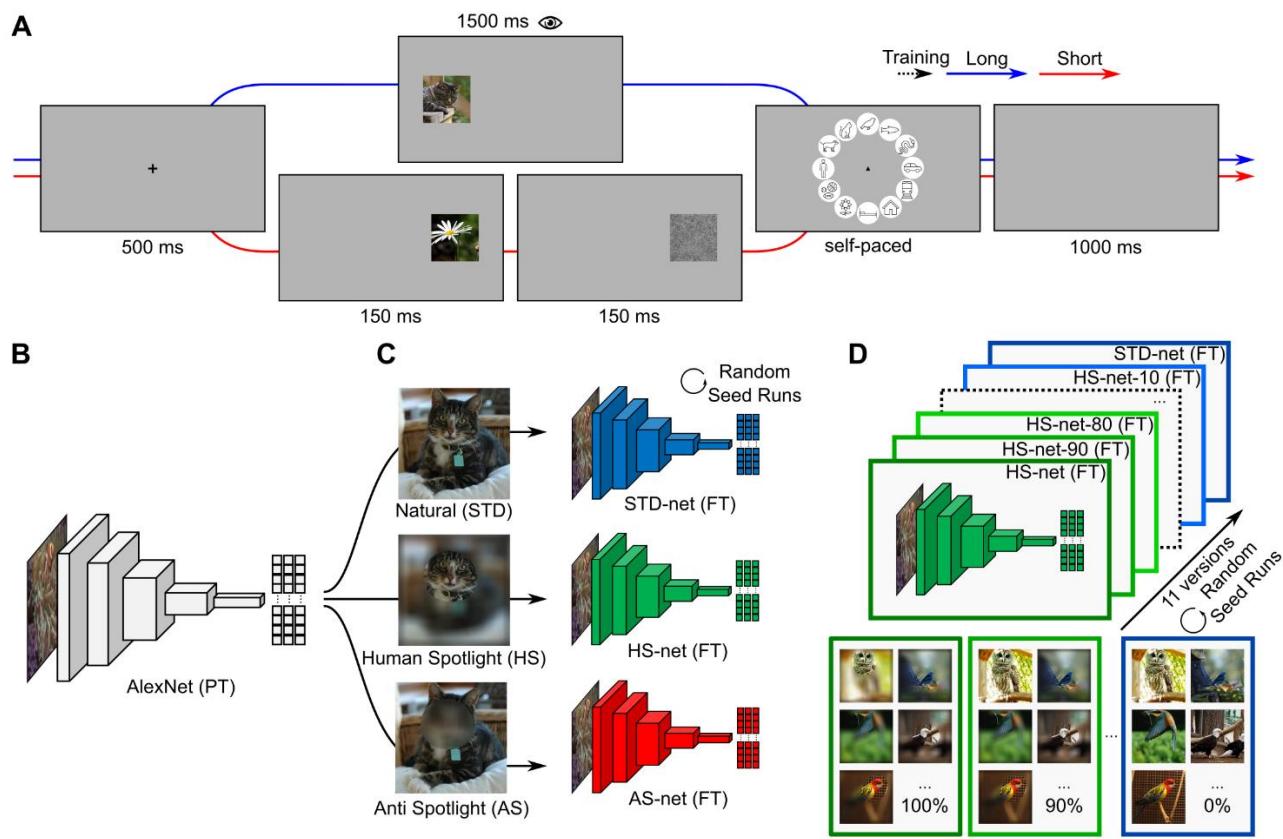

**Figure 1.** Conceptual overview. **(A)** Experimental design of the eye tracking paradigm. The experiment consists of two blocks. Trials in the first block consist of a fixation cross (500 ms), followed by an image presented on either the left or the right side for a long duration (1500 ms), a circular categorization screen with category icons (self-paced with a time limit of 5000 ms), and a blank interstimulus interval (1000 ms). Trials in the second block are similar except that an image was presented only briefly (150 ms) and followed by a visual backward mask (1/f pink noise; 150 ms). **(B)**

Pretrained AlexNet (PT) consists of an image input, five convolutional, and three fully connected layers. **(C)** AlexNet versions are fine-tuned (FT) on identical images of twelve categories with different data-driven blurring manipulations (STD-nets = standard, HS-nets = Human-Spotlight, AS-nets = Anti-Spotlight). **(D)** HS-nets and AS-nets with different manipulation ratios in the training dataset are fine-tuned ranging from all HS and AS images (100 %) to all STD images (0 %) in steps of 10 %. The same procedure, here visualized for HS-nets, was applied to AS-nets with the corresponding manipulation type.

## 2  Methods

### 2.1  General Procedure

For the comparison of object recognition in humans and DCNNs, we attempted to create an experimental setup considering literature on fair and insightful human-machine comparisons (Firestone, 2020; Funke et al., 2020). Therefore, we adopted various measures to align the tasks accordingly. A few examples that aim to restrict certain one-sided advantages are the consistent use of entry-level categories, a forced-choice categorization, and an image presentation of 150 ms followed by a visual backward mask for the evaluated categorization data. First, the eye tracking experiment was conducted. Here, data from block one (i.e., long image presentation of 1500 ms) were used to compute heatmaps and to enable a comparison of several time windows across the course of feedforward and recurrent processing, while data from block two (i.e., short image presentation of 150 ms) were devoted to behavioral measures such as accuracy and error patterns based on feedforward visual processing only. The eye tracking data from one half of the images were used to create image versions with different guided focus manipulations. These images were then used to fine-tune three different AlexNet manipulation types (hereafter *standard AlexNet* = STD-net, *Human-Spotlight AlexNet* = HS-net, *Anti-Spotlight AlexNet* = AS-net) with otherwise identical training procedures. The only difference here were the differently blurred versions of the same fine-tuning examples. The obtained models were then tested on the other half of the images. Both eye tracking heatmaps and GradCAMs were also used from incorrectly classified images (while GradCAMs were always extracted for the true category label) to avoid comparisons contingent on subjective responses (VanRullen, 2011).

### 2.2  Human Observers – Eye Tracking Experiment

In the eye tracking experiment, a total of 49 participants (25 female, 22 male, 2 other) with an age ranging from 18 to 62 years (M = 24.60, SD = 8.07) were tested. Participants had normal or corrected-to-normal vision, no problems with color vision, and no other eye diseases. One participant had to be excluded from the analyses due to inaccurate eye tracking calibration. The experimental procedure was admitted by the University of Salzburg ethics committee, in line with the declaration of Helsinki, and agreed to by participants via written consent prior to the experiment. Psychology students were compensated for taking part in the study with accredited participation hours.

**Guiding Visual Attention in DCNNs**

The experimental paradigm (see Figure 1A) consisted of two blocks with 360 trials each. During trials of the first block with long image presentation, participants were presented a fixation cross for 500 ms, followed by an image positioned at ~13 degrees of visual angle on either the left or right side for 1500 ms. The original image was scaled by factor two (454-by-454 px and 11.52 degrees of visual angle wide), while the original resolution (227-by-227 px) was kept. After that, a categorization screen with circularly arranged icons of the twelve entry-level categories was displayed for a maximum duration of 5000 ms. Participants were instructed to choose a category as fast as possible with a mouse click. The origin of the mouse cursor was set to the center of the circle in each trial. The order of category icons was fixed with roughly animate categories on the top and inanimate categories on the bottom half to aid memorability of icon locations. A blank screen with an interstimulus interval of 1000 ms terminated the trial. Trials in the second block with short image presentation were identical except for a brief image presentation of only 150 ms, which was followed by a visual backward mask (1/f pink noise) for the same duration. All participants had to complete twelve practice trials before starting the actual experiment. At the end of the practice trials, open questions could be discussed and clarified. Training trials were excluded from analyses. The image datasets of the first and second block with 360 images each were interchanged across participants, the order of the presented images within blocks was fully randomized for each participant, and the assignment of image datasets to blocks as well as the presented image position were randomized across different versions of the experiment. With this design, it was possible to simultaneously obtain eye tracking data during longer time periods of image presentation (i.e., 1500 ms), to test the similarity between eye tracking heatmaps and GradCAMs across time, as well as categorizations based on brief image presentations (i.e., 150 ms), to compare categorizations as a result of assumed feedforward visual processing for every image across two participants. The experiment took approximately one hour to complete and was divided by short breaks in-between and within blocks. The experiment was programmed with ExperimentBuilder (SR Research Ltd., Mississauga, ON, Canada) and conducted in a controlled laboratory setting. Participants were seated in front of a screen (1920-by-1080 px, 60 Hz) and had to place their head onto a chin rest at a distance of 60 cm. An EyeLink 1000 (SR Research Ltd., Mississauga, ON, Canada) desktop mount was used to track the participants' right eye at a sampling rate of 1000 Hz. Recorded eye movements were preprocessed and analyzed in MATLAB (Version 2022a, The MathWorks, Inc., Natick, Massachusetts, United States) using the Edf2Mat Toolbox (Etter & Biedermann, 2018). First, all timepoints with x- and y-coordinates within the presented image coordinates were grouped into time windows of 50 ms before heatmaps were compiled for these time windows of shifting onset. In this way, the first 1000 ms within the image boundaries were analyzed, allowing a maximum of 20 discrete time windows. After that, the obtained heatmaps were downscaled to original image size (227-by-227 px) and a Gaussian filter with a standard deviation of 20 px was applied. Lastly, heatmaps of individual participants were averaged and normalized per image.

## 2.3   DCNNs – Fine-Tuning and Saliency Maps

The investigated models were trained using MATLAB and Deep Learning Toolbox (Release 2022a, The MathWorks, Inc., Natick, Massachusetts, United States) using a conventional transfer learning approach. Here, an implementation of AlexNet (see Figure 1B; Krizhevsky et al., 2012), consisting of an image input, five convolutional, as well as three fully connected layers, and pretrained on >1 million images of 1000 categories from the ImageNet database (Deng et al., 2009), was fine-tuned on the images from the publicly available Ecoset dataset (Mehrer et al., 2021), which were also presented in the first block of the eye tracking experiment. Unmanipulated, STD-images were used for fine-tuning STD-net, while manipulated HS-images and AS-images were used for HS-nets and AS-nets, respectively (see Figure 1C). During the fine-tuning procedure, the last fully connected, softmax, and classification layer were replaced while earlier layers were frozen and the models retrained with the Stochastic Gradient Descent with Momentum (SGDM) optimizer, a mini-batch size of 42, an initial learn rate of $10^{-4}$, and a weight learn rate factor of 20 in mentioned layers. All models were trained with varying image versions but identical settings for a maximum of 30 epochs and a validation patience of five epochs. No data augmentation was applied throughout the process. Saliency maps were computed based on the models last convolutional layer using the GradCAM attribution algorithm (Selvaraju et al., 2017) for the true class label. The last convolutional layer marks the final processing stage of the encoding procedure, contains information used for subsequent classification, and in turn corresponds best to the theoretical concept of eye movements as a final processing stage and behavioral result of visual perception. In this way, the saliency map (227-by-227 px) represents the normalized contribution of a specific pixel to the class probability score of the true image category. Unlike eye tracking heatmaps, saliency maps do not vary across time for static images. The previously reported grid-phenomena of saliency map activations (van Dyck et al., 2021) should not influence the comparability in this study, as all models share identical architecture.

Additionally, to investigate the proposed accuracy versus similarity trade-off of our guided focus approach, HS-nets with different ratios of HS- and STD-images within categories were fine-tuned (see Figure 1D). These models ranged from versions fine-tuned on only manipulated images (HS-net-100) to versions trained on only STD-images (HS-net-0) with steps of 10 % in-between. This resulted in eleven different versions, where for example three images per category were unmanipulated in the HS-net-90, half of the images were unmanipulated in the HS-net-50, and all images were manipulated and unmanipulated in the HS-net-100 (i.e., HS-net) and HS-net-0 (i.e., STD-net), respectively.

To increase the statistical power of both main analyses and overcome the general weakness of commonly used fixed-value comparisons, we estimate the models' accuracy and similarity by repeating this process with random initialization seeds (e.g., Mehrer et al., 2021) as well as shuffled training and validation splits for the number of compared human participants (N = 23) and ten times per ratio in the accuracy versus similarity trade-off analysis. Hence, we train and test the model throughout different runs and average all obtained values to improve signal-to-noise ratio and further reduce the influence of individual unmanipulated or manipulated images.

## 2.4 Image Dataset

Data for fine-tuning the DCNNs consisted of images from the Ecoset testing set (Mehrer et al., 2021). In the HS condition, images which were displayed to human participants for 1500 ms during the first experimental block were modified in the way that image parts with fixation densities below the first tertile were blurred and above remained unchanged to create a blurred out surrounding of the focal points of fixations. The inverse procedure was used in the AS condition. For blurring, a Gaussian filter (W = 30, SD = 7) was applied to the image followed by another Gaussian filter as an edge taper (W = 35, SD = 9). For the eye tracking experiment and DCNN testing, a dataset of 360 images in total, consisting of twelve entry-level categories, with six animate categories (namely *human, dog, cat, bird, fish,* and *snake*), six inanimate categories (namely *car, train, house, bed, flower,* and *ball*), and 30 images each was used. All images were randomly drawn and preprocessed in identical procedure. Images were cropped to the largest central square and resized to fit AlexNet's image input layer (227-by-227 px). A manual visual inspection ensured that images did not include multiple categories of interest, overlaid text, and image effects or did not show the object anymore due to cropping. All violations were removed. Furthermore, to identify face presence as well as object and face regions of interest (ROI), two independent raters manually detected and segmented the images of the testing dataset with in-house built MATLAB scripts. The obtained object ROIs (N = 360) and face ROIs (N = 133) were averaged, normalized, and a Gaussian filter was applied (W = 10, SD = 5, see Supplementary Figures). To investigate the detection of face ROIs, a face detection index was computed through the proportion of visual attention within compared to outside of the circumscribed region. This resulted in high scores in cases where visual attention was focused specifically on face regions and lower scores when attention was distributed across the image.

# 3 Results

## 3.1 Performance

To investigate object recognition performance, categorization data of human participants acquired under short image presentation of 150 ms were compared to the testing results of DCNN runs. A Welch-ANOVA, due to violated homoscedasticity (indicated by significant Levene's tests), revealed significant differences between the groups [$F(3,48.44) = 147.78$, $p < .001$, $\eta^2 = .77$; see Figure 2]. Bonferroni-corrected pairwise t-tests revealed significant differences between human participants (M = 89.80, SD = 3.18) and standardly fine-tuned STD-nets [M = 88.27, SD = 1.06; $t(30.91) = 2.31$, $p = .028$, $d = 0.64$]. Nevertheless, this significant but marginal accuracy difference demonstrates that DCNNs should principally be able to reach human benchmark performance with the fine-tuning and experimental procedures at hand. As expected, both HS-nets (M = 82.22%, SD = 1.92) and AS-nets (M = 81.11, SD = 1.57) were significantly outperformed by human participants [HS-net: $t(41.81) = 10.21$, $p < .001$, $d = 28.21$; AS-net: $t(37.28) = 12.33$, $p < .001$, $d = 28.21$] and STD-nets [HS-nets: $t(33.89) = 13.26$, $p < .001$, $d = 3.91$; AS-nets: $t(38.29) = 18.25$, $p < .001$, $d = 5.38$] due to noise added by both guided focus manipulation types. However, contrary to our expectations, AS-nets did not significantly exceed HS-nets even though in the AS manipulation images were blurred to a lesser extent (M = 23.62, SD = 0.06) compared to the HS manipulation (M = 76.38, SD = 0.06). Instead, the opposite was observed, as HS-nets even significantly outperformed AS-nets [t(42.26) = 2.15, p = .038, d = 0.63]. Moreover, further exploratory analyses of accuracies across categories and common misclassifications indicated unique behavioral patterns (see Supplementary Figures).

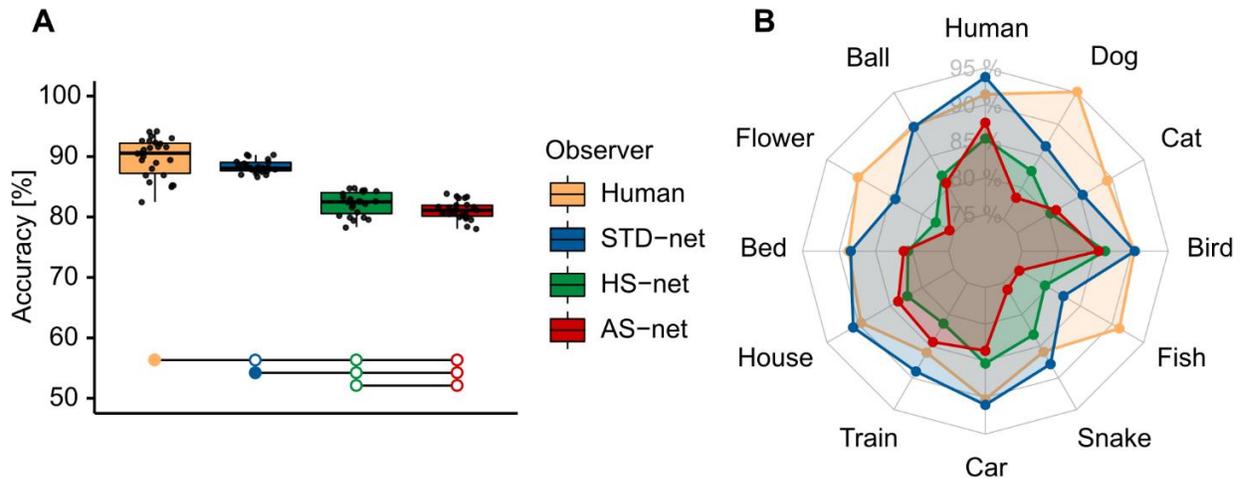

**Figure 2** | Mean object recognition accuracy of human participants and DCNN runs. **(A)** Mean accuracy of human participants, standardly fine-tuned STD-nets, purposefully manipulated HS-nets, and inversely manipulated AS-nets across all testing images. Groups marked by filled dots outperformed groups marked by empty dots significantly at the p < .05 level. **(B)** Mean accuracy profiles across categories.

## 3.2 Visual Attention

Average human eye tracking heatmaps acquired under long image presentation of 1500 ms and DCNN GradCAMs were compared to test the human-likeness of individual models (for examples see Supplementary Figures). Therefore, mean Pearson correlation coefficients between eye tracking heatmaps and GradCAMs for individual testing images were transformed into Fisher's-Z-scores ($Z_r$) to aid linear comparability (see Figure 3). A vast distribution of scores was observed. Due to violated normal distribution of the data (indicated by significant Shapiro-Wilk tests), the following statistical analyses were performed nonparametrically. For two-factorial nonparametric tests, we use an advanced statistical procedure from the *nparLD* R software package and report its ANOVA-type statistic (ATS; Noguchi et al., 2012). Moreover, as pairwise correlations coefficients between eye tracking heatmaps and different GradCAMs could be considered as differences within the same group (i.e., human participants) and thereby may result in dependencies, we test the coefficients with overlap using the *cocor* R package (Diedenhofen & Musch, 2015) and also report the Hittner et al. (2003) modification of Dunn and Clark's (1969) approach of a back-transformed average Fisher's Z procedure.

**Guiding Visual Attention in DCNNs**

As hypothesized, a Kruskal-Wallis test indicated significant differences between the three models [$H(2) = 26.52$, $p < .001$, $\eta^2 = 0.02$]. Generally, in terms of our guided focus manipulation, we expected HS-nets to demonstrate more human-like visual attention compared to their inversely conceptualized AS-net counterparts and commonly fine-tuned STD-nets. Most interestingly, Bonferroni-corrected pairwise Wilcoxon tests indeed revealed that AS-nets (Mdn = 0.10) focused on significantly less human-like image parts compared to STD-nets (Mdn = 0.24; W = 77332, $p < .001$, r = 0.17; z = 4.07, $p < .001$) and HS-nets (Mdn = 0.23, W = 77149, $p < .001$, r = .16; z = 3.23, $p < .001$). However, no overall significant differences between HS-nets and STD-nets were found (W = 64527, p = .922, r = .003; z = -0.10, p = .459). Remarkably, as hypothesized, a significant interaction effect between manipulation type and animacy was identified [$F_{ATS}(1.62) = 10.26$, $p < .001$] and as the overall patterns seemed to remain stable across animacy groups, an additional main effect of animacy may be assumed [$F_{ATS}(1) = 26.09$, $p < .001$]. Subsequent Bonferroni-corrected pairwise Wilcoxon tests revealed significant differences for both animate and inanimate objects between AS-nets compared to STD-nets (animate: W = 20543, $p < .001$, r = .23; z = 3.60, $p < .001$ / inanimate: W = 18353, p = .029, r = .11; z = 2.04, p = .021) and HS-nets (animate: W = 20507, $p < .001$, r = .23; z = 2.89, p = .002 / inanimate: W = 18288, p = .035, r = .11; z = 1.61, p = .054). Generally speaking, most human-like visual attention was obtained with STD-nets and HS-nets categorizing animate objects, while AS-nets categorizing inanimate objects elicited the least human-like attentional processes. Finally, to further pinpoint the central effects of our manipulation, the influence of face presence (i.e., human or animal faces) on the similarity of visual attention was analyzed. The analysis revealed a significant interaction effect between manipulation type and face presence [$F_{ATS}(1.56) = 7.48$, p = .002] with similar patterns to the reported animacy splits. Here again, in both face and nonface images, AS-nets focused on significantly more dissimilar image parts compared to STD-nets (face: W = 11211, $p < .001$, r = .23; z = z = 3.14, $p < .001$ / nonface: W = 30004, p = .002, r = .14; z = 2.67, p = .004) and HS-nets (face: W = 11456, $p < .001$, r = .26; z = 2.78, p = .003 / nonface: W = 29364, p = .010, r = .12; z = 1.89, p = .029). Taken together, these results imply that the distinct biases for living entities and especially their faces in human gaze patterns may have introduced the opposite effects in AS-nets. As an interim summary, it should be noted that the guided focus approach did not seem to increase the overall fit to human eye tracking data as intended in GradCAMs of HS-nets. However, the single most striking observation to emerge from the data was that HS-nets and AS-nets did show only a marginal accuracy difference and yet the AS manipulation lead to significantly less human-like visual attention. Besides that, it should be noted that individual categories were investigated through exploratory analyses only and no statistical procedures were applied due to the reported accuracy differences.

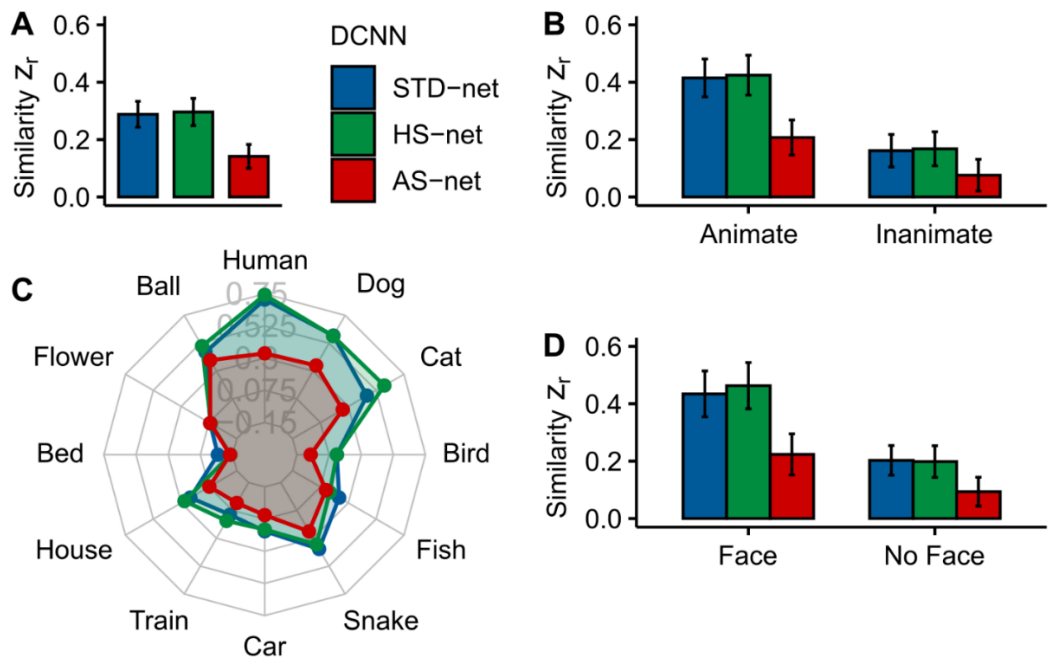

**Figure 3** | Mean Fisher's Z transformed Pearson correlations between DCNN GradCAMs and human eye tracking heatmaps for all individual testing images. **(A)** Overall similarity of standardly fine-tuned STD-nets, purposefully manipulated HS-nets, and inversely manipulated AS-nets compared to eye tracking heatmaps. **(B)** Similarity across animacy. **(C)** Similarity across categories. **(D)** Similarity across face presence. Error bars display bootstrap confidence intervals.

## 3.3   Face Detection

Following up on the observed effects of face presence, especially in AS-nets, we compared the models' GradCAMs to face ROIs, which were visually segmented by two independent raters. Therefore, a face detection index (see Methods Section) was computed. While data from human participants displayed a strong face bias (Mdn = 2.74), which varied considerably across time, DCNNs generally demonstrated clearly less focused attention towards face regions (STD-nets: Mdn = 0.59; HS-nets: Mdn = 0.57; AS-nets: Mdn = 0.42, see Figure 4). A Kruskal-Wallis test indicated a significant effect between the models [H(2) = 7.54, p = .023, $\eta^2$ = .01] and ensuing Bonferroni-corrected pairwise Wilcoxon tests further revealed significant differences between AS-nets compared to STD-nets (W = 10351, p = .016, r = .15) and HS-nets (W = 10319, p = .019, r = .14). These results highlight the importance of faces in human vision, as our AS manipulation seemed to robustly blur them from the fine-tuning dataset and AS-nets thereby shifted their visual attention away from them to classify images without losses in accuracy compared to HS-nets.

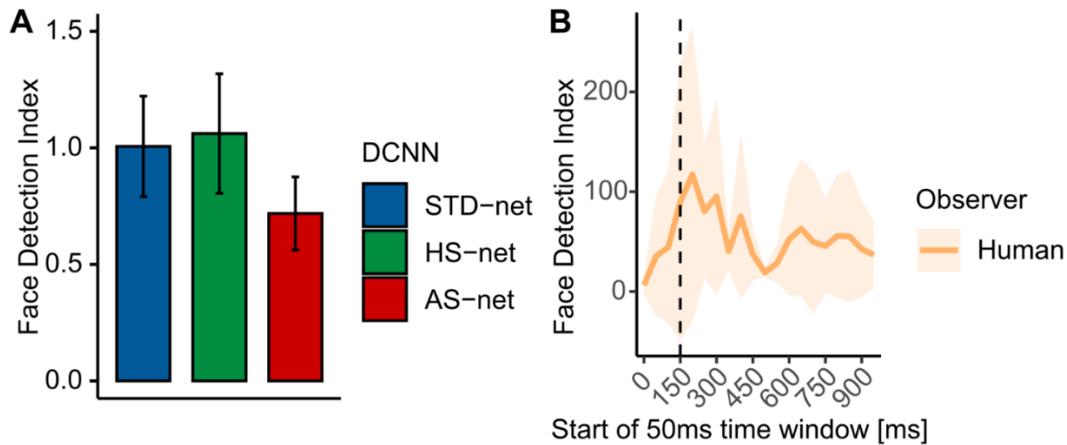

**Figure 4** | Mean face detection indices. The face detection index reflects the proportional attention within compared to outside of the face ROI. **(A)** Mean face detection indices for GradCAMs overall. **(B)** Mean face detection index for eye tracking heatmaps across time. **Note:** Shaded areas display bootstrap confidence intervals. The dotted line marks the transition from feedforward to recurrent processing at ~150ms.

## 3.4 Visual Attention Across Time

In a next step, we included the time course of human object recognition processes in our comparison. We hypothesized that during the first feedforward sweep, preattentive processing should prevail before attentive processing can set in and human observers saccade to relevant object features. Hence, in terms of the fit of the computational models, we expected the similarity of visual attention to increase rapidly and peak after this initial time window (i.e., due to similar bottom-up processes) and then gradually diminish due to the onset of increasingly feedback processes (i.e., missing top-down processes in the models). The data resembled our proposed time course only partly, as indeed the similarity increased reliably at ~150-200 ms but seemed to reach a plateau without a noticeable decrease in later time windows (see Figure 5). Splits by animacy and face presence revealed similar patterns to the overall analysis while the previously implied but not significant difference between HS-nets and STD-nets in face images seemed to emerge at the beginning of attentive processing. Moreover, analyses by category revealed remarkably specific time courses, which seem to be the source of the vast distribution of similarity scores and should be treated with caution due to the reported accuracy differences.

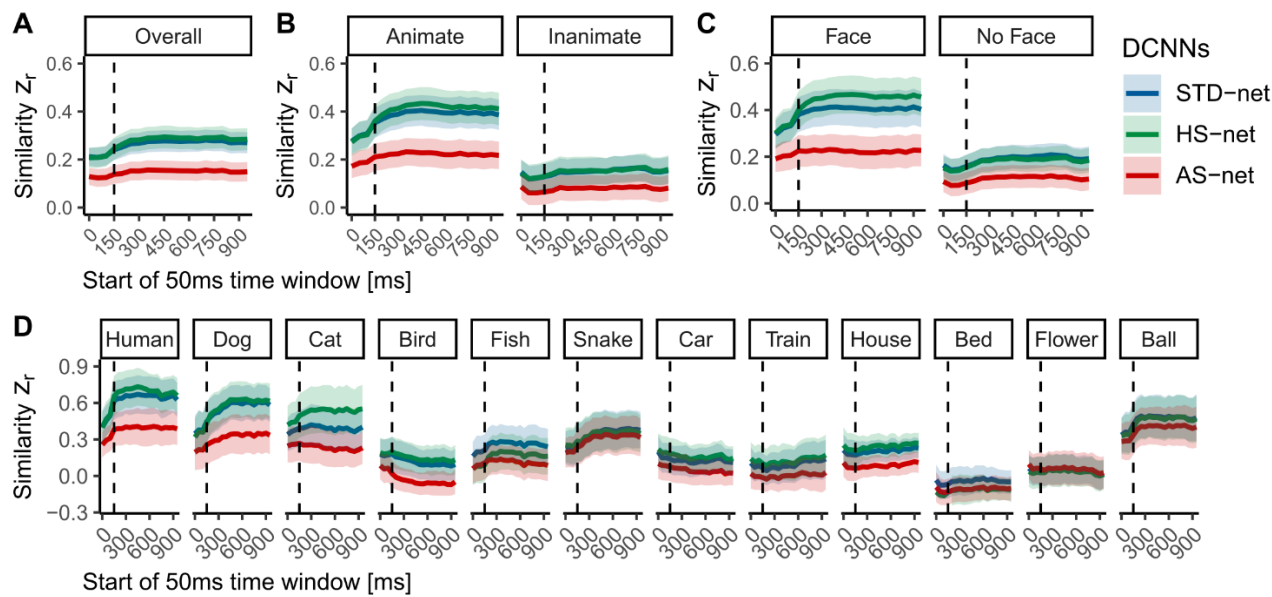

**Figure 5** | Mean Fisher's Z transformed Pearson correlations between DCNN GradCAMs and eye tracking heatmaps for all individual testing images across time. **(A)** Overall similarity of standardly fine-tuned STD-nets, purposefully manipulated HS-nets, and inversely manipulated AS-nets compared to eye tracking heatmaps across the first 1000 ms within the image boundaries. **(B)** Similarity across time split by animacy. **(C)** Similarity across time split by face presence. **(D)** Similarity across time split by category. Shaded areas display bootstrap confidence intervals. The dotted line marks the transition from feedforward to recurrent processing at ~150ms.

## 3.5   Accuracy-Similarity Trade-Off

Finally, to investigate possible methodological advantages and disadvantages of the applied approach, we performed accuracy and similarity analyses for ratios of HS and AS manipulations (see Methods Section). We hypothesized that accuracy and similarity change as a function of magnitude in respect to the directionality of the manipulation as smaller manipulation ratios were expected to return both measurements to the STD-net baseline. Hence, while in HS-nets, smaller manipulation ratios (i.e., more STD-images in the fine-tuning dataset) were expected to lead to increases in accuracy but decreases in human-likeness, in AS-nets both accuracy and similarity were expected to increase. The results endorsed this assumption partially (see Figure 6), as AS-nets but not HS-nets followed this pattern. Taken together, these findings endorse the idea that the guided focus approach influences the human-likeness of the models at least in the negative direction considerably and with reasonable losses in accuracy.

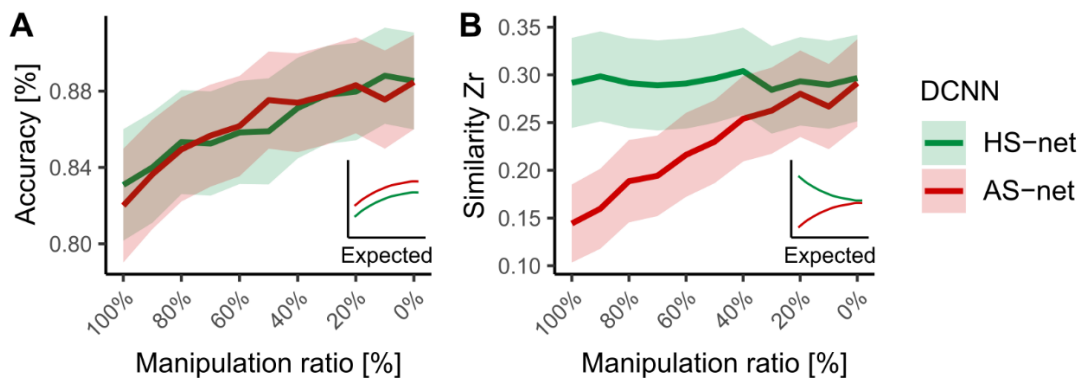

**Figure 6** | Accuracy versus similarity trade-off across HS and AS manipulation ratios ranging from 100 % to 0 % in steps of 10 %. **(A)** Mean object recognition accuracy of DCNN runs for all individual testing images. **(B)** Mean Fisher's Z transformed Pearson correlations between DCNN GradCAMs and human eye tracking heatmaps for all individual testing images. Shaded areas display bootstrap confidence intervals.

# 4 Discussion

In this study, we address the question whether it is possible to influence the human-likeness of DCNNs through a purely data-driven approach. In two opposing guided focus manipulations (i.e., HS-nets and AS-nets), we used eye tracking data to selectively modify the informational content in the training examples and thereby attempted to shift the models' visual attention either towards or away from human-relevant features. Subsequently, we compared and validated the resulting models against new eye movement data.

Generally, analyses of object recognition accuracy implied that human benchmark performance, often reported for the applied architecture (Krizhevsky et al., 2012), was nearly reached by the standardly fine-tuned models (i.e. STD-nets). As anticipated, the guided focus manipulation led to significant losses in accuracy (i.e., in HS-nets and AS-nets) due to the overall reduction of informational content. While there were no significant accuracy differences between HS-nets and AS-nets, this finding turned out to be highly intriguing in the light of the proportional extent of the manipulations and the associated differences in visual attention. Even though in the HS manipulation more than three times as many features were blurred as in the AS manipulation, both models indicated similar accuracy. The HS manipulation was quantitatively more interfering but spared qualitatively more important features. This may illustrate the interplay between object-relevant features and contextual shortcut learning in DCNNs (for review see Geirhos et al., 2020), as both processes represent valid strategies and yet reflect the distribution of information across natural images (for review see Oliva & Torralba, 2007) through their unequal efficiency. Furthermore, the resulting similarities between GradCAMs and eye tracking heatmaps demonstrated significant differences in human-likeness among the models. Contrary to our expectations, these findings did not show a significantly increased overall human-likeness in the experimental condition (i.e., HS-nets). Nevertheless, the hypothesized effect was present in the control condition (i.e., AS-nets), as these models focused significantly more on dissimilar image parts compared to humans. Across following analyses, remarkable results were uncovered.

Firstly, the overall similarity for individual images indicated category-specific effects. This is in accordance with our previous findings on comparisons of human and DCNN visual attention (van Dyck et al., 2021) and may reflect the diverse processes captured by human eye movements (for review see Schütz et al., 2011) as well as their transferability to DCNNs through the applied approach. Here, it is important to note that, in the eye tracking experiment, we cannot determine the exact timepoint at which exogenous, recognition-relevant neural processes are completed and overt visual attention may be influenced by other endogenous, recognition-irrelevant neural processes. This way, it is entirely possible that the overlap between recognition-relevant and recognition-irrelevant but salient features may be highly variable across categories. To further investigate this, we analyzed the human-likeness of GradCAMs across different time windows. The results support the notion, that during feedforward, preattentive processing (i.e., ~150-200 ms within the image boundaries) subsequent feedback, attentive processing (i.e., saccades and fixations) may be programmed. However, the fit to our computational models only partly resembled the expected time course, as the similarity rapidly increased at ~150 ms but did not pursue the proposed decline, due to increasingly recurrent/feedback processes that lack in the models. It seems likely that the expected time course and especially the decrease in similarity later on were not observable for specific categories (e.g., *house*, *bed*, and *flower*), as their image examples may display predominantly non-critical, distributed features (e.g., textural surfaces) and not critical, focal features (e.g., shape configurations) as other categories (e.g., *human*, *dog*, and *cat*), which would attract fixations more reliably. Naturally, it is difficult to investigate the impact of feedforward and recurrent processing on eye movements as they may be a mere behavioral proxy of the underlying neural processes (Van der Stigchel et al., 2009). On the one hand, while the initial dip, interpreted as preattentive processing, could be an artifact of the central fixation bias (Tatler, 2007), both interpretations are not necessarily contradicting. On the other hand, as objects are usually, and in our case, located in the center of natural images (i.e., due to "photographer bias", see Supplementary Figures), this should have led to higher similarity right from the beginning.

Secondly, we were able to show that animate objects and face presence both significantly increase the effect of our manipulation. These findings are in line with previous results on visual processing of living beings, as they are known to elicit more stereotypical gaze patterns with more frequent and longer fixations compared to inanimate objects (Jackson & Calvillo, 2013; Kovic et al., 2009; New et al., 2007; Yang et al., 2012). Following analyses of images including faces helped to further isolate the presumed driving force of our manipulation, as AS-nets detected significantly less face regions compared to other models. Our results corroborate a large body of literature on the unique attentional (Buswell, 1935; Crouzet et al., 2010; Gilchrist & Proske, 2006; Tatler et al., 2010) and computational (Farah et al., 1998; Haxby et al., 2000; Kanwisher, 2000; Kanwisher et al., 1997) status of faces in human vision as a basis for automatic and rapid attraction of fixations. We believe that through this "face bias", found in the eye tracking data, AS-nets learned to selectively ignore face configurations as informative features for object recognition. It remains debatable whether this presumably acquired "face blindness" may indeed support the face expertise hypothesis (Gauthier & Bukach, 2007; Gauthier & Nelson, 2001) and disagree with the innate face bias (Johnson et al., 1991; Johnson & Mareschal, 2001). Interestingly, while Xu et al. (2021) reported the emergence of a face module in fully face-deprived AlexNets, Blauch et al. (2021) discovered noticeable effects of expertise on face recognition in DCNNs that also support our findings. This way, the AS manipulation may not only be useful to further investigate this debate but could also offer plausible computational models for disrupted face perception in disorders such as autism spectrum disorder and prosopagnosia.

We are aware of the possibility that the guided focus manipulation induces merely cosmetic effects. This would explain the absent increased human-likeness in the HS manipulation, as DCNNs may not be able to process the features emphasized by human eye movements adequately. Therefore, future investigations on the interaction effect of biologically inspired architectures and human-relevant data will be needed. However, we believe that the human-likeness of DCNNs may not yet be directly linked to their performance, as manifest, accuracy-relevant, and latent, similarity-relevant information does not necessarily have to be identical. This further strengthens our confidence that the manipulation should principally work in both directions but is challenging to implement and verify. Thus, due to numerous limitations, the effect may have been solely found in the less challenging direction. While saliency maps were criticized to be highly noisy (Kim et al., 2019), our averaging approach should have increased the overall signal-to-noise ratio but also might have amplified the gravitational impact of the previously reported grid-phenomena in saliency maps towards certain points. Furthermore, our approach may be considerably underpowered, as fine-tuning influences especially the last fully connected layer, but saliency maps are extracted from the minimally reweighted last convolutional layer. While we still found the reported differences in human-likeness, future work will benefit from larger eye tracking datasets and fully trained models. From a neuroscience perspective, a visual search paradigm or controversial stimuli (Golan et al., 2020) with different foci of bottom-up and top-down visual attention may be better suited to disentangle the role of feedforward and recurrent processing in the human-likeness of DCNNs. However, they are challenging to implement in object recognition of mutually exclusive targets. From a computer science perspective, future work on recurrent architectures (Mnih et al., 2014) and soft spatial attention (Lindsay, 2020), reweighting solely the representations and not directly entire image parts, may be able to further push the boundaries of such a data-driven approach to design more human-like models without the reported disadvantages. Our findings could have several possible implications for computer vision in the form of explainable artificial intelligence. While architectural modifications may help to better understand which mechanisms a given model uses to come to a general decision (i.e., global interpretability), data-driven manipulations may be able to shed light on the mechanisms that underlie a single decision (i.e., local interpretability; Adadi & Berrada, 2018). We hope that our novel approach excites both fields and sparks innovative ideas for future computational models of vision.

# 5  Acknowledgements

We thank our colleagues Charlotte Paulina Schöllkopf and Rade Kutil for providing insightful comments in our lively discussions.

# 6  Author Contributions

LD, SD, and WG contributed to the conception and design of the study. LD and SD programmed the experiment. LD wrote the first draft of the manuscript. LD and SD collected the eye tracking data. LD implemented the computational models. LD and WG analyzed and interpreted the data. All authors contributed to manuscript revision, read, and approved the submitted version.

# Supplementary Figures

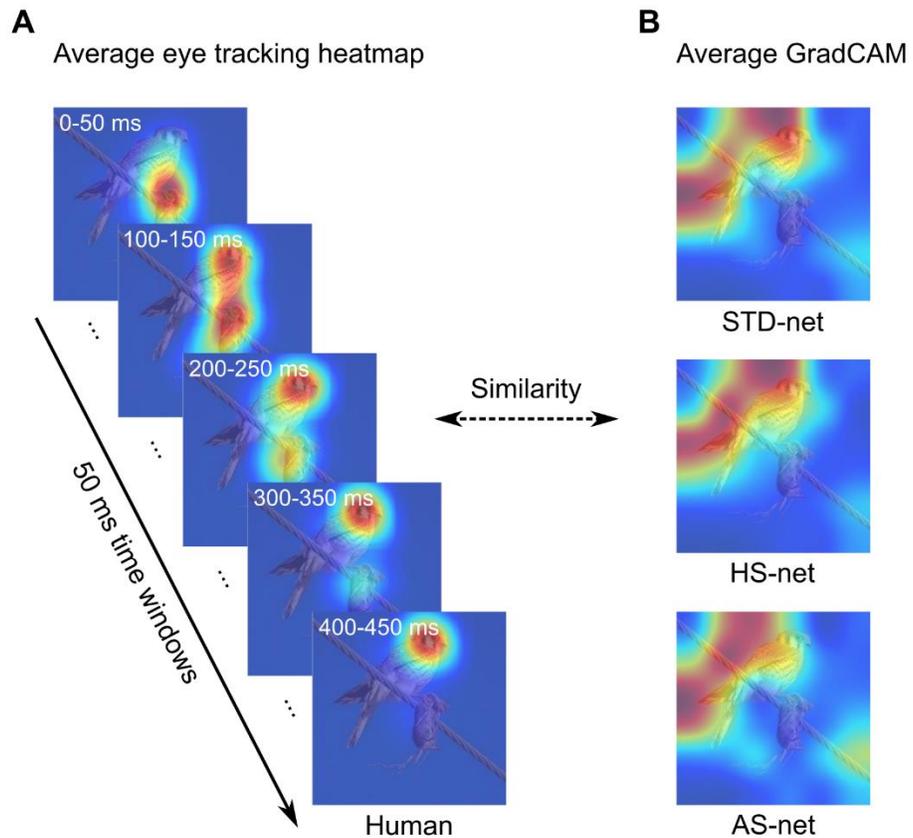

**Supplementary Figure 1** | Image example with average visual attention of human participants and DCNN manipulation types. (**A**) Average human eye tracking heatmaps across 50 ms time windows. (**B**) Average DCNN GradCAM saliency maps for individual manipulation types.

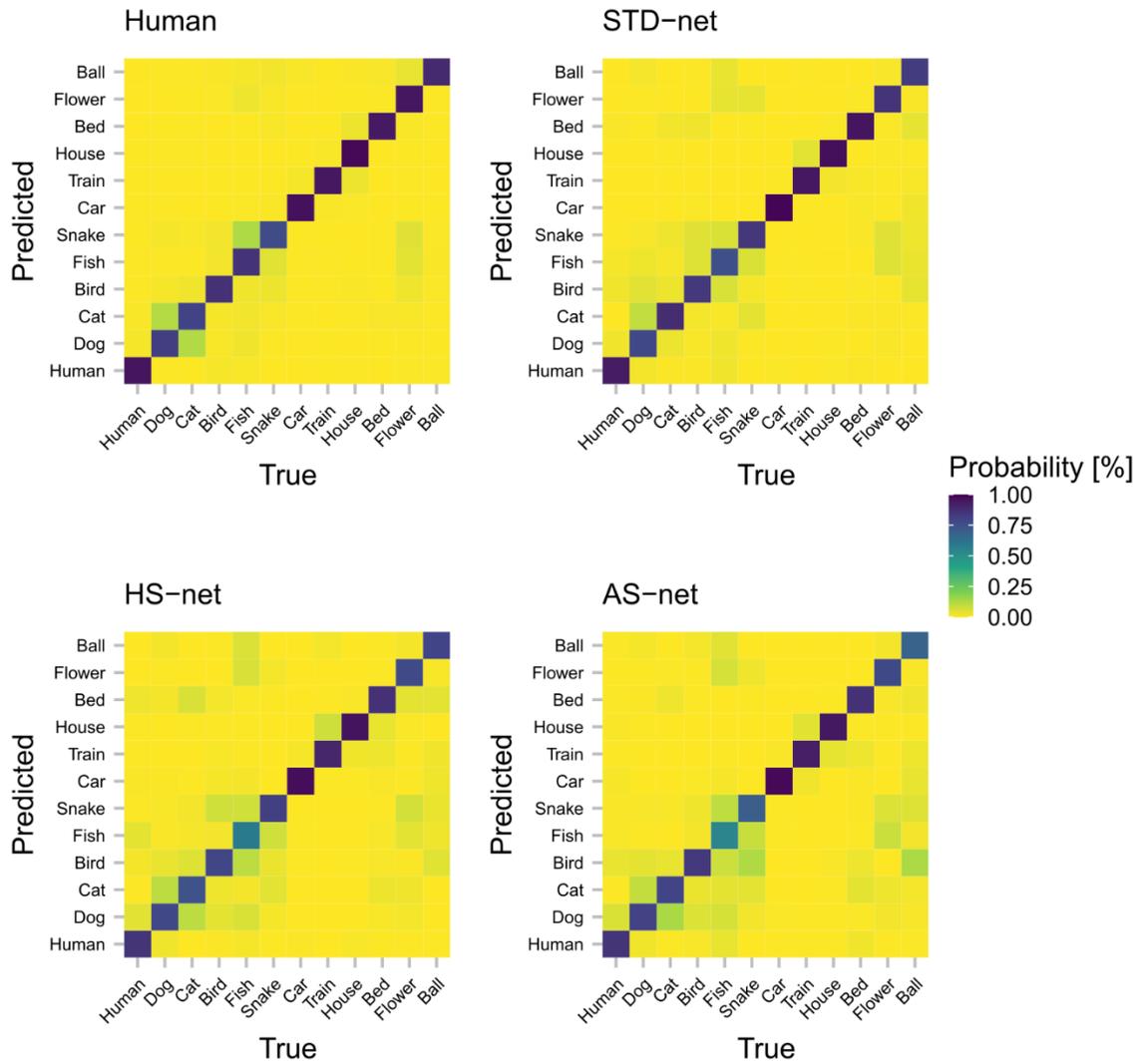

**Supplementary Figure 2** | Confusion matrices displaying common miscategorizations by probability of occurrence for true versus predicted category combinations in human participants, standardly fine-tuned STD-nets, purposefully manipulated HS-nets, and inversely manipulated AS-nets across all testing images.

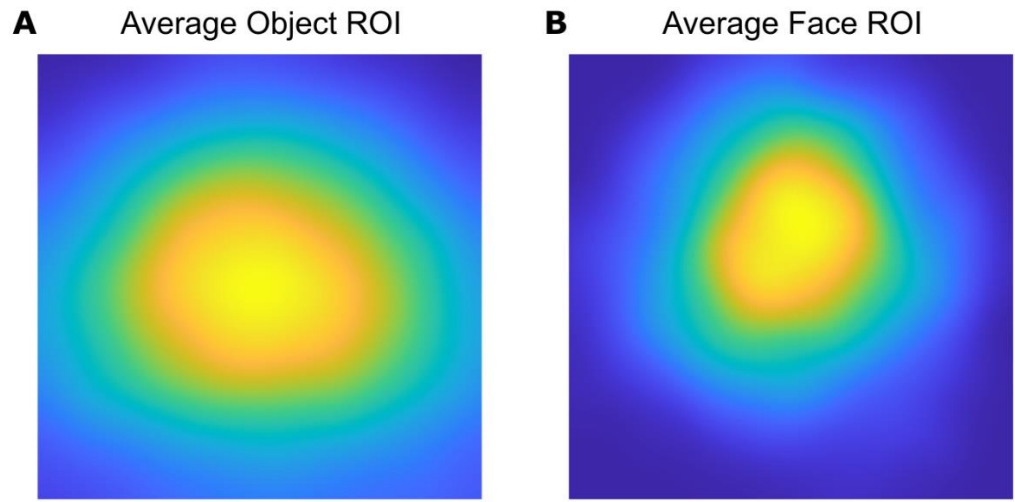

**Supplementary Figure 3** | Average manually segmented regions of interest (ROIs). **(A)** Average object ROI (N = 360). **(B)** Average face ROI (N = 133).